\documentclass[11pt,a4paper]{article}
\usepackage[utf8]{inputenc}
\usepackage{naaclhlt2018}
\usepackage{times}
\usepackage{latexsym}
\usepackage{paralist}
\usepackage{url}
\usepackage{subcaption}
\usepackage{booktabs}
\aclfinalcopy 

\title{Experiments with Universal CEFR Classification}

\author{Sowmya Vajjala \\
  Applied Linguistics and Technology Program \\
 Iowa State University, USA \\
  {\tt sowmya@iastate.edu} \\
  \And
  Taraka Rama \\
  Department of Informatics \\
  University of Oslo, Norway \\
  {\tt tarakark@ifi.uio.no} \\
  }

\date{}

\hypersetup{draft}
\begin{document}
\maketitle
\begin{abstract}
The Common European Framework of Reference (CEFR) guidelines describe language proficiency of learners on a scale of 6 levels. While the description of CEFR guidelines is generic across languages, the development of automated proficiency classification systems for different languages follow different approaches. In this paper, we explore universal CEFR classification using  domain-specific and domain-agnostic, theory-guided as well as data-driven features. We report the results of our preliminary experiments in monolingual, cross-lingual, and multilingual classification with three languages: German, Czech, and Italian. Our results show that both monolingual and multilingual models achieve similar performance, and cross-lingual classification yields lower, but comparable results to monolingual classification.
\end{abstract}

\section{Introduction}
Automated Essay Scoring (AES) refers to the task of automatically grading student essays written in response to some prompt. Different approaches for AES have been proposed in literature, where it is modeled as a regression, ranking or a classification problem \citep[cf.][]{Yannakoudakis.Briscoe.ea-11,Taghipour.Ng-16,Pilan.Alfter.ea-16}. To our knowledge, all the previous work described approaches that work with a single language (mostly English). Feature representations that can work for multiple languages and those that support cross-lingual AES have not been explored. 

At first thought, using an essay scoring model developed for one language to test on another language seems unacceptable. However, CEFR guidelines are not developed for a specific language. This leads us to hypothesize about a common model of ``proficiency'' that can work across languages. The existence of such a model would also be beneficial for quick prototyping of AES systems for languages that do not have readily available training data.

In this paper, we explore this hypothesis by exploring CEFR-classification for three languages- German, Italian, and Czech, for which CEFR graded data is publicly available. Apart from constructing individual models using generic text classification and AES specific features, we also looked into cross-lingual (i.e., training a model on one language and testing on another) and multilingual classification approaches (i.e., building a single classification model trained on all the three languages at once). 

Testing our universal CEFR hypothesis would require a common feature representation across languages. We developed such a representation, by employing features based on part-of-speech tags and dependency relations from the Universal Dependencies (UD)\citep{Nivre.deMarneffe.ea-16} project which provides treebanks for over 60 languages.\footnote{\url{http://universaldependencies.org/}} Therefore, this approach can be easily extended to other languages with available CEFR graded texts and UD treebanks.

In short, the contributions of this paper are as follows: 
\begin{enumerate}
\item We study AES for multiple languages for the \emph{first} time using CEFR scale.
\item We explore, for the \emph{first} time, the possibility of a Universal CEFR classifier by training a single model consisting of three languages.
\item We also report \emph{first} results on cross-lingual AES.
\end{enumerate}

The rest of this paper is organized as follows: Section~\ref{sec:related} describes related work. Section~\ref{sec:approach} describes our data and methods. Section~\ref{sec:results} discuss our experiments and results in detail. Section~\ref{sec:conclusion} concludes the paper with pointers to future work.

\section{Related Work}
\label{sec:related}
AES is a well studied research problem and AES systems are used to automatically grade essays in exams such as GRE\textsuperscript{\textregistered} and TOEFL\textsuperscript{\textregistered} \cite{Attali.Bustein-04}. There is a considerable amount of work that explored various aspects of AES research such as: dataset development, feature engineering, multi-corpus studies and the role of prompt and task information \cite{Yannakoudakis.Briscoe.ea-11,Phandi.Chai.ea-15,Zesch.Wojatzki.ea-15,Alikaniotis.Yannakoudakis.ea-16,Taghipour.Ng-16,Vajjala-18}.

AES models developed for non-English languages, primarily using the CEFR scale (\citealt{Hancke-13} for German, \citealt{Pilan.Alfter.ea-16} for Swedish, \citealt{Vajjala.Loo-14} for Estonian) employ several language specific features and show their relevance for the task. However, to the best of our knowledge, there is no previous work on developing common models and feature representations that work across languages. Against this background, we set out to address the question: ``Is there a universal model for language proficiency classification?'' 

\section{Approach} 
\label{sec:approach}
\subsection{Dataset}
To test our hypotheses, we need corpora graded with CEFR scale for multiple languages. One such multi-lingual corpus is the freely available MERLIN \cite{Boyd.Hana.ea-14} corpus.\footnote{\url{http://merlin-platform.eu/}} This corpus consists of 2286 manually graded texts written by second language learners of German (DE), Italian (IT), and Czech (CZ) as a part of written examinations at authorized test institutions. The aim of these examinations is to test the knowledge of the learners on the CEFR scale which consists of six categories -- A1, A2, B1, B2, C1, C2 -- which indicate improving language abilities. The writing tasks primarily consisted of writing formal/informal letters/emails and essays. MERLIN corpus has a multi-dimensional annotation of language proficiency covering aspects such as grammatical accuracy, vocabulary range, socio-linguistic awareness etc., and we used the ``Overall CEFR rating'' as the label for our experiments in this paper. Other information provided about the authors included- age, gender, and native language, and information about the task such as topic, and the CEFR level of the test itself. We did not use these information in the experiments reported in this paper. Further, we removed all Language-CEFR Category combinations that had less than 10 examples in the corpus (German had 5 examples for level C2 and Italian had 2 examples for B2 which were removed from the data). We also removed all the unrated texts from the original corpus. The final corpus had 2266 documents covering three languages, and Table~\ref{tab:corpus} shows the distribution of labels in the final corpus. 

\begin{table}[htb!]
\begin{center}
\small
\begin{tabular}{lccc}
\toprule \bf CEFR level & \bf DE & \bf IT & \bf CZ \\ \midrule
A1 &57 & 29& 0\\
A2 &306 &381 & 188 \\
B1 & 331&393 & 165\\
B2 & 293& 0& 81\\
C1 & 42& 0&0 \\
\midrule
Total & 1029&803 &434 \\ \bottomrule
\end{tabular}
\end{center}
\caption{Composition of MERLIN Corpus}
\label{tab:corpus} 
\end{table}

\subsection{Features}
Our feature set consists of features that are commonly used in AES systems, as well as others that can be generalized across languages. They are described below:
\begin{enumerate}
\item Word and POS n-grams, which were commonly used in AES models in the past \cite{Yannakoudakis.Briscoe.ea-11}.
\item Task-specific word and character embeddings trained through a softmax layer. Although word embeddings were used in recent neural AES models\cite{Alikaniotis.Yannakoudakis.ea-16}, this paper is the first to explore character embeddings as a cross-linguistic feature for AES model. 
\item Dependency n-grams where each unigram is a triplet consisting of dependency relation, POS tag of the dependent, POS tag of the head. To our knowledge, these features were not used in any of the previous work on AES.
\item Linguistic features specific to AES literature:
\begin{enumerate}
\item Document length: The number of words in a document which is a common feature used in AES literature.
\item Lexical richness features: \citet{Lu-12} described several lexical richness and language proficiency for English, which were used in previous AES systems \citep{Hancke-13}. In this paper, we used lexical density, lexical variation, and lexical diversity features that are commonly used in the AES literature.
\item Error features: Total number of errors and total spelling errors are obtained for German and Italian from an open-source, rule based spelling and grammar checker.\footnote{\url{https://languagetool.org/}} To the best of our knowledge, there is no existing tool for Czech grammar check, and hence we did not extract error features for Czech.
\end{enumerate}
We will refer to these as domain features in this paper. 
\end{enumerate}

We extracted all n-gram features where $n\in [1,5]$ and excluded those n-grams that appeared less than 10 times in the corpus. All the POS and dependency relation based features are extracted using the UDPipe parser \cite{Straka.Hajic.ea-16} trained on Universal Dependencies treebanks \citep{Nivre.deMarneffe.ea-16}.

\paragraph{Feature Combinations:} In addition to the above mentioned features, we also explored the effectiveness of combining n-gram features with domain features. The n-gram features are sparse whereas the domain features are dense; therefore, we combined them by training a n-gram feature classifier and using the probability distribution over its cross-validated predictions with domain features to train the final classifier. 

\subsection{Classification and Evaluation}
We compared logistic regression, random forests, multi-layer perceptron, and support vector machines for experiments with non-embedding features and Neural Network models trained on task-specific embedding representations for other experiments. Word embeddings for each language were task-specific are trained only using the MERLIN corpus. The embeddings are stacked with a softmax layer and trained with categorical cross-entropy loss and Adadelta algorithm. We also experimented by training a softmax classifier with character and word embeddings as input and found that the combined model does not perform as well as a stand-alone word embeddings model.

Considering the space restrictions, we report only the best performing systems in this paper. Due to the unbalanced class distribution across all the three languages in the data, we employed weighted-F1 score to evaluate the performance of our trained models. Weighted F1 is computed as the weighted average of the F1 score for each label, taking label support (i.e., number of instances for each label in the data) into account. For both monolingual and multilingual settings, we report results with 10-fold cross validation. For cross-lingual evaluation, we report results on the test language's data.

All our neural network models are implemented using Keras \citep{chollet2015keras} with TensorFlow as the backend \citep{tensorflow2015-whitepaper} and other models were implemented using scikit-learn \cite{scikit-learn,sklearn-api}.\footnote{Relevant code, generated results and the parameter settings are available at: \url{https://github.com/nishkalavallabhi/UniversalCEFRScoring}}

While it is also possible to model AES as a regression task, we report classification results which is common in CEFR classification tasks. Our initial experiments with linear regression gave Pearson and Spearman correlation in the range of $0.7-0.9$ with gold standard scores, which is comparable with previous results on English AES task obtained using regression models \cite{Alikaniotis.Yannakoudakis.ea-16}.

\section{Experiments and Results}\label{sec:results}

For all the experiments, we considered a classifier using only document length (number of words per document) as the feature as the baseline. Unless explicitly stated, all the reported results for non-embedding features are based on Random Forest classifier, which was the best performing classifier in our experiments. Numbers with superscript \textsuperscript{L} indicate performance of results with a Logistic Regression model.

\subsection{Monolingual classification}
Our classification results with different feature sets for the three languages are summarized in table~\ref{tab:monoresults}.

\begin{table}[htb!]
\begin{center}
\begin{tabular}{lccc}
\toprule \textbf{Features} & \textbf{DE} & \textbf{IT} & \textbf{CZ} \\
\midrule 
Baseline &0.497 & 0.578\textsuperscript{L} & 0.587\textsuperscript{L} \\
 Word ngrams (1) &0.666&0.827 & 0.721 \\
 POS ngrams (2) & 0.663& 0.825& 0.699\\
 Dep. ngrams (3) & 0.663& 0.813&0.704 \\
 Domain features &0.533\textsuperscript{L} & 0.653\textsuperscript{L}& 0.663\\\midrule
 (1) + Domain &\textbf{0.686} &\textbf{0.837} &\textbf{0.734} \\
 (2) + Domain & \textbf{0.686}& 0.816& 0.709\\
 (3) + Domain &0.682& 0.806& 0.712\\
\midrule Word embeddings &0.646 & 0.794& 0.625\\ \bottomrule
\end{tabular}
\end{center}
\caption{Weighted F1 scores for Monolingual Classification}
\label{tab:monoresults} 
\end{table}

All feature representations perform better than the document length baseline, resulting in close to 25\% improvement in the macro F1 score in some cases. All the three sets of n-gram features perform comparably in the case of German and Italian. In the case of Czech, word n-grams turn out be a better predictor of CEFR scale than syntactic features. The domain features, by themselves, do not perform well for any of the languages. However, concatenating the domain features with n-gram features yield slightly better classification results. Word embeddings perform poorly for Czech compared to other non-embedding features, and come close to lexical and syntactic features in the case of German and Italian. Whether using embeddings pre-trained on  a larger corpus will give us better scores is something that needs to be explored in future.

To our knowledge, \citet{Hancke-13} is the only comparable work which explored CEFR classification for German using the same dataset, but with several language specific morphological and syntactic features. Our results are comparable to the reported results of \citet{Hancke-13}, although we primarily rely on data-driven features. To our knowledge, there are no existing results for Czech and Italian. 

German, which has a larger dataset, seems to perform poorer than the other two languages. One possible explanation for this could be that we are dealing with a 5 class classification for German, where as it is only a 3 class problem for Czech and Italian. It is also possible that these feature representations are not sufficient to model German language proficiency labeling task. Further experiments (and possibly with other existing CEFR datasets) are needed to understand why the classification results differ between different languages. 

\subsection{Multilingual classification}
In this setup, we combined all the language texts and trained a single universal CEFR classifier. Table~\ref{tab:multiresults} shows the results. For the non-neural models, we experimented with and without considering language information as a categorical feature. The neural network model is a multitasking model \citep{ccoltekin2016discriminating} that consists of character and word embeddings as input. The model learns to predict both the language of the text (language identification) and the CEFR category simultaneously. The model is trained using categorical cross-entropy and Adadelta algorithm. The table shows results with and without language identification for neural models. 

\begin{table}[htb!]
\begin{center}
\begin{tabular}{lcc}
\toprule \textbf{Features} & \textbf{lang (-)} & \textbf{lang (+)} \\
\midrule 
Baseline &0.428\textsuperscript{L} & - \\
Word n-grams & 0.721 & 0.719\\
POS n-grams & \textbf{0.726} & \textbf{0.724}\\
Dependency n-grams &0.703 & 0.693\\
Domain features & 0.449\textsuperscript{L}& 0.471\textsuperscript{L}\\\midrule
Word + Char embeddings & 0.693 & 0.689 \\ \bottomrule 
\end{tabular}
\end{center}
\caption{Weighted F1 scores for multilingual classification with models trained on combined datasets.}
\label{tab:multiresults}
\end{table}

We observe that the document length baseline seems to perform poorer than monolingual models in this case. Further, we can see that the average result on monolingual model as close to the multilingual model in case of POS n-grams, dependency n-grams, and embeddings. However, domain features clearly perform poorly compared to monolingual case. While one could argue that the better performance multilingual model over some monolingual models is due to more training data, this does not seem to be true for some feature groups (baseline, domain features). One inference we can draw is that some feature groups have similarities in terms of proficiency categories assigned for different languages, which lends support to our hypothesis. Although we did not perform a qualitative language specific evaluation yet, the results so far indicate that efforts to build such a universal scoring model is a worthwhile effort. 

\subsection{Cross-lingual classification}
In this setup, we trained a CEFR model on one language and tested it on others. We trained the cross-lingual model only on German data since it has examples for all categories in our corpus. Table~\ref{tab:crossresults} summarizes our results. We did not train with word n-grams and word embeddings here as they are lexical and are language specific and are not suitable for this scenario. Table~\ref{tab:crossresults} presents the results of the experiments in this setup.
\begin{table}[htb!]
\begin{center}
\begin{tabular}{lcc}
\toprule \textbf{Features} & \textbf{Test:IT} & \textbf{Test:CZ} \\
\midrule 
Baseline & 0.553\textsuperscript{L} &0.487\textsuperscript{L}\\
POS n-grams & \textbf{0.758}& 0.649\\
Dependency n-grams & 0.624& \textbf{0.653}\\
Domain features &0.63\textsuperscript{L} & 0.475\\ \bottomrule
\end{tabular}
\end{center}
\caption{Weighted F1 scores for cross-lingual classification model trained on German.}
\label{tab:crossresults} 
\end{table}
The results show a drop in performance when compared to monolingual models, which is not surprising as the feature weights are tuned to German syntactic features. However, it is interesting to note that the drop is less than 10\% in both cases. In the case of Italian, the domain features yield similar results to monolingual results suggesting that there are some possible universal patterns of language use in the progression towards language proficiency. All feature groups perform better than the document length baseline for Italian, and domain features perform poorer than the baseline for Czech. The confusion matrices for these experiments (cf. tables~\ref{tab:crossconf1} and \ref{tab:crossconf2}) suggest that most of the misclassification occurs only between adjacent levels of proficiency.

\begin{table}[htb!]
\begin{subtable}{\linewidth}\centering{
\begin{tabular}{l l l l l l} \toprule
$\rightarrow$ Pred & A1 &A2&B1&B2&C1 \\
\midrule A1 & 5& 24 &0 & 0& 0\\
 A2 & 9 & 311 & 56 & 5 &0\\
 B1 & 1 & 70 & 279 & 44 &0\\ \bottomrule 
\end{tabular}
}
\caption{DE-Train:IT-Test setup with POS n-gram features}
\label{tab:crossconf1} 
\end{subtable} \newline\newline\newline\newline
\begin{subtable}{\linewidth}\centering{
\begin{tabular}{l l l l l l} \toprule
$\rightarrow$ Pred & A1 &A2&B1&B2&C1 \\
\midrule A2 & 0 & 129 & 57 & 2 &0\\
 B1 & 0 & 23 & 101 & 41 &0\\ 
 B2 & 0 & 5 & 25 & 51 & 0 \\ \bottomrule 
\end{tabular}
}
\caption{DE-Train:CZ-Test setup with Dependency features}
\label{tab:crossconf2} 
\end{subtable}
\caption{Confusion matrices for cross-lingual scoring with Random Forests by training on German data (DE-train).}
\label{tab:crossconf}
\end{table}

The results of this experiment indicate that while cross-lingual classification results in a drop in performance, it still captures the proficiency scale meaningfully. So, the next step in this direction would be to explore better representations of the data, and  better modeling methods. 

\section{Conclusion}
\label{sec:conclusion}
In this paper, we reported the results of first experiments conducted with the aim of exploring a ``universal CEFR classifier''. The results so far indicate that cross-lingual and multilingual classifiers yield comparable performance to individual language models. These results provide some evidence for a universal notion of language proficiency and leave open many questions which need to be explored further in future. Our immediate future plans include a systematic exploration of feature representations which are meaningful for the AES context while being portable across languages. Modeling proficiency classification as a domain adaptation problem (where the domain is another language), and doing multi-task learning by considering other annotation dimensions are other interesting directions to pursue in future. Considering that we have publicly available CEFR graded corpora for other languages such as Estonian, it would be interesting to extend this approach to new languages. This would enable us to investigate questions such as the relationship between genetic/typological similarities between languages and cross/multi-lingual CEFR classification task in future. 

When it comes to using such methods in real world language testing applications, researchers express concerns about the validity of the chosen feature constructs, and bias and fairness in models. Some recent research \cite{Madnani.Loukina.ea-17} in this direction leaves us with some pointers to incorporate these aspects in future research.

\section*{Acknowledgments}
The second author is supported by BIGMED\footnote{\url{https://bigmed.no}}, a Norwegian Research Council funded Lighthouse project, which is gratefully acknowledged.

\bibliography{naaclhlt2018}
\bibliographystyle{acl_natbib}
\appendix
\section{Supplemental Material}
\label{sec:supplemental}
The code and data relevant for our experiments are available at: \url{https://zenodo.org/badge/latestdoi/108113378}.

\end{document}